%% file: arxiv_v2.tex
\documentclass[10pt,twocolumn,letterpaper]{article}

\usepackage{cvpr}              


\definecolor{cvprblue}{rgb}{0.21,0.49,0.74}
\usepackage[pagebackref,breaklinks,colorlinks,allcolors=cvprblue]{hyperref}
\usepackage{amsmath}
\usepackage{amssymb}   
\usepackage{bm}
\usepackage{algorithmic}
\usepackage{algorithm}
\usepackage{graphicx}
\usepackage{array}
\usepackage{makecell}
\usepackage{multirow}
\usepackage{hhline}

\newcommand{\PreserveBackslash}[1]{\let\temp=\\#1\let\\=\temp}
\newcolumntype{C}[1]{>{\PreserveBackslash\centering}p{#1}}
\newcolumntype{R}[1]{>{\PreserveBackslash\raggedleft}p{#1}}
\newcolumntype{L}[1]{>{\PreserveBackslash\raggedright}p{#1}}

\def\paperID{1679} 
\def\confName{CVPR}
\def\confYear{2025}

\title{Arbitrary-steps Image Super-resolution via Diffusion Inversion\vspace{-1mm}}

\author{Zongsheng Yue$^{1,2}$, ~ Kang Liao$^{2}$,~ Chen Change Loy$^{2}$ \vspace{2mm} \\ 
$^{1}$Xi'an Jiaotong University, Xi'an, China \\
$^{2}$S-Lab, Nanyang Technological University, Singapore \\
}

\begin{document}

\twocolumn[{%
\renewcommand\twocolumn[1][]{#1}%
\maketitle
\vspace{-12mm}
\begin{center}
    \centering
    \includegraphics[width=0.97\linewidth]{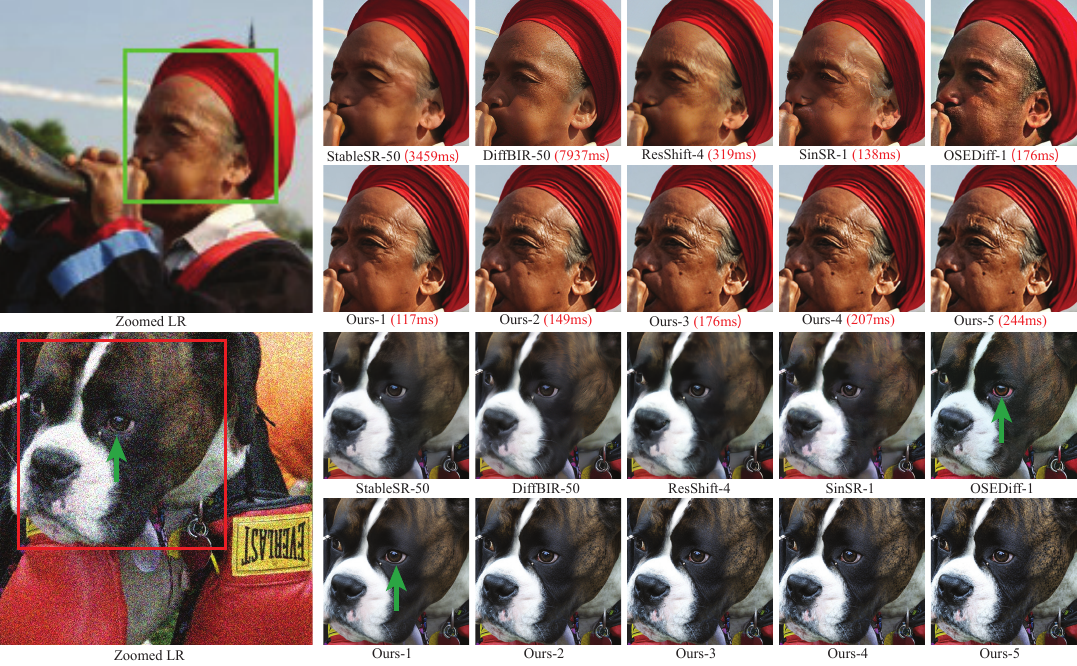}
    \vspace{-2mm}
    \captionof{figure}{
    Qualitative comparisons of our proposed method to recent state-of-the-art diffusion-based approaches on two real-world examples, where the number of sampling steps is annotated in the format ``Method name-Steps''. We provide the runtime (in milliseconds) highlighted by \textcolor{red}{red} in the sub-caption of the first example , which is tested on $\times$4 ($128\rightarrow 512$) SR task on an A100 GPU. Our method offers an efficient and flexible sampling mechanism, allowing users to freely adjust the number of sampling steps based on the degradation type or their specific requirements. In the first example, mainly degraded by blurriness, multi-step sampling is preferable to single-step sampling as it progressively recovers finer details. Conversely, in the second example with severe noise, a single sampling step is sufficient to achieve satisfactory results, whereas additional steps may amplify the noise and introduce unwanted artifacts. \textit{(Zoom-in for best view)}
    } 
    \label{fig:teaser}
\end{center}%
}]

\begin{abstract}
This study presents a new image super-resolution (SR) technique based on diffusion inversion, aiming at harnessing the rich image priors encapsulated in large pre-trained diffusion models to improve SR performance. We design a \textit{Partial noise Prediction} strategy to construct an intermediate state of the diffusion model, which serves as the starting sampling point. Central to our approach is a deep noise predictor to estimate the optimal noise maps for the forward diffusion process. Once trained, this noise predictor can be used to initialize the sampling process partially along the diffusion trajectory, generating the desirable high-resolution result. Compared to existing approaches, our method offers a flexible and efficient sampling mechanism that supports an arbitrary number of sampling steps, ranging from one to five. Even with a single sampling step, our method demonstrates superior or comparable performance to recent state-of-the-art approaches. The code and model are publicly available at \url{https://github.com/zsyOAOA/InvSR}.\vspace{-2mm}
\end{abstract}

\section{Introduction}
Image super-resolution (SR) is a fundamental yet challenging problem in computer vision, aiming to restore a high-resolution (HR) image from a given low-resolution (LR) observation. The main challenge of SR arises from the complexity and often unknown nature of the degradation model in real-world scenarios, making SR an ill-posed problem. Recent breakthroughs in diffusion models~\cite{sohl2015deep,ho2020denoising,song2021scorebased}, particularly large-scale text-to-image (T2I) models, have demonstrated remarkable success in generating high-quality images. Owing to the strong generative capability of these T2I models, recent studies have begun to use them as a reliable prior to alleviate the ill-posedness of SR. This work follows this research line, further exploring the potential of diffusion priors in SR.

The prevailing SR approaches leveraging diffusion priors usually attempt to modify the intermediate features of the diffusion network, either through optimization~\cite{kawar2022denoising,chung2022come,wang2023zeroshot} or fine-tuning~\cite{wang2023zeroshot,lin2023diffbir,yang2023pasd,wu2024seesr}, to better align them with the given LQ observations. 
In this work, we propose a new technique based on diffusion inversion to harness diffusion priors. Unlike existing approaches, it attempts to find an optimal noise map as the input of the diffusion model, without any modification to the diffusion network itself, thereby maximizing the utility of diffusion prior. 

While considerable advances have been made in generative adversarial networks (GANs)~\cite{goodfellow2014generative} inversion for various applications~\cite{zhu2020domain,xia2022gan}, including SR~\cite{gu2020image,pan2021exploiting,chan2022glean}, extending these principles to diffusion models presents unique challenges, particularly for SR tasks that demand high fidelity preservation. 
In particular, the multi-step stochastic sampling process of diffusion models makes inversion non-trivial. The straightforward inversion approach to optimize the distinct noise maps at each diffusion step is expensive and complex.
Additionally, the iterative inference mechanism would accumulate prediction errors and randomness at each step, which can significantly compromise fidelity. Therefore, recent diffusion inversion methods have mainly focused on tasks with lower fidelity requirements, such as image editing~\cite{mokady2023null,garibi2024renoise}. 

In this work, we reformulate diffusion inversion for the more challenging task of SR.
To enable diffusion inversion for SR, we introduce a deep neural network called noise predictor to estimate the noise map from a given LR image. 
In addition, a Partial noise Prediction (PnP) strategy is devised to construct an intermediate state for the diffusion model, serving as the starting point for sampling.
%
This is made possible by \textit{adding noise onto the LR image} according to the diffusion model's forward process, \textit{where the noise predictor predicts the added noise instead of random sampling}. 
This approach is driven by the following key motivations:
\begin{itemize}
    \item \textbf{Rationality}. 
    LR and HR images differ only in high-frequency details. With the addition of appropriate noise, the LR image becomes indistinguishable from its HR counterpart. Thus, the noisy LR can serve as a proxy for deriving the inversion trajectory during reverse diffusion.
    %
    \item \textbf{Complexity}. Rather than predicting noise maps for all diffusion steps, the PnP strategy simplifies the inversion task by limiting predictions to the starting step, thereby reducing the overall complexity of the inversion process. 
    \item \textbf{Flexibility}. The noise predictor can be trained to predict noise maps for multiple predefined starting steps. During inference, we can freely select a starting step from them and then use any existing sampling algorithm with an arbitrary number of steps, offering favorable flexibility in controlling the sampling process. 
    \item \textbf{Fidelity}. The starting steps during training are carefully selected to have a high signal-to-noise ratio (SNR), ensuring robust fidelity preservation for SR. In practice, we enforce an SNR threshold greater than 1.44, corresponding to the timestep of 250 in Stable Diffusion~\cite{rombach2022high}. 
    \item \textbf{Efficiency}. As the sampling process begins from a step earlier than 250 (SNR larger than 1.44), the PnP strategy effectively \textit{reduces the number of sampling steps to fewer than five} when combined with off-the-shelf accelerated sampling algorithms~\cite{song2021denoising,karras2022elucidating}. This addresses the common inefficiency issue in diffusion-based SR approaches.
\end{itemize}

Unlike most existing diffusion-based methods that rely on fixed sampling steps, our flexible sampling mechanism offers a versatile solution for handling varying degrees of degradation in SR. In SR, it is common to encounter different types and intensities of corruption. Intuitively, the number of sampling steps should adapt to the specific degradation conditions. For example, as shown in Fig.~\ref{fig:teaser}, multi-step sampling is preferable to single-step sampling in the first case, as it effectively reduces blurriness and restores finer details. In contrast, for the second example with severe noise, a single sampling step achieves satisfactory results, while additional steps may amplify the noise and introduce unwanted artifacts. Our method uniquely allows users to adjust sampling to suit different degradation types.

The main contributions of this work are twofold. First, we propose a novel SR approach based on diffusion inversion, which effectively leverages the diffusion prior by integrating an auxiliary noise predictor while keeping the entire diffusion backbone fixed. Second, our method introduces a flexible and efficient sampling mechanism that allows for arbitrary sampling steps, ranging from one to five. Remarkably, even when the steps are reduced to just one, our approach still achieves superior or comparable performance to recent dedicated one-step diffusion methods.

\section{Related Work}

\noindent\textbf{Diffusion Prior for SR.}
Existing diffusion prior-based SR approaches can be broadly categorized into two classes. The first class of methods involves re-optimizing the intermediate results of the diffusion model to ensure consistency with the given LR images via pre-defined or estimated degradation models. Representative works include DDRM~\cite{kawar2022denoising}, CCDF~\cite{chung2022come}, and DDNM~\cite{wang2023zeroshot}, among others~\cite{chung2022improving,fei2023generative,chung2023diffusion,song2023pseudoinverseguided,murata2023gibbsddrm,yue2024difface,xiao2024dreamclean}. While effective, these methods are limited by their computational complexity, as they require solving an optimization problem at each diffusion step, leading to slow inference. Furthermore, they often rely on manually defined assumed degradation models and thus cannot handle the blind SR problem in real-world scenarios. The second class directly fine-tunes a pre-trained large T2I model for the SR task. StableSR~\cite{wang2024exploiting} pioneers this paradigm by incorporating spatial feature transform layers~\cite{wang2018recovering} to guide the T2I model toward generating HR outputs. Subsequent works follow by proposing various fine-tuning strategies to exploit diffusion priors, including DiffBIR~\cite{lin2023diffbir}, SeeSR~\cite{wu2024seesr}, PASD~\cite{yang2023pasd}, S3Diff~\cite{zhang2024degradation}, and so on~\cite{xia2023diffir,yu2024scaling,sun2024coser,wu2024one,li2024distillation,noroozi2024you}. These methods have achieved impressive performance, validating the effectiveness of diffusion priors for SR. 

\vspace{1mm}\noindent\textbf{Diffusion Inversion.} 
Diffusion inversion focuses on determining the optimal noise map set that, when processed through the diffusion model, reconstructs a given image.
DDIM~\cite{song2021denoising} first addressed this by generalizing the diffusion model via a class of non-Markovian processes, thereby establishing a deterministic generation process. Subsequent approaches, such as those by Rinon~\etal~\cite{gal2022image} and Mokady~\etal~\cite{mokady2023null}, proposed optimizing the text embedding to better align with the desired textual guidance. Recent efforts have further refined the optimization strategies for both the textual and visual prompts~\cite{miyake2023negative,nguyen2024visual}, as well as for intermediate noise maps~\cite{wallace2023edict,meiri2023fixed,garibi2024renoise,zhang2024exact,kang2024eta,ju2024pnp}, leading to notable enhancements in inversion quality. Despite these advances, existing methods mainly focus on image editing and cannot meet the high-fidelity requirements of SR.

In this work, we tailor the diffusion inversion technique for SR. While Chihaoui~\etal~\cite{chihaoui2024blind} have recently explored diffusion inversion for image restoration, their method relies on solving an optimization problem at each inversion step, significantly limiting its inference efficiency. In contrast, our approach introduces a noise prediction module that, once trained, enables efficient inversion without requiring iterative optimization during inference. This leads to substantial improvements in both the efficiency and practicality of diffusion inversion for SR tasks.  

\section{Methodology}
In this section, we present the proposed diffusion inversion technique for SR. To maintain consistency with the notations used in diffusion models, we denote the LR image as $\bm{y}_0$ and the corresponding HR image as $\bm{x}_0$. 

\subsection{Motivation}
The diffusion model~\cite{sohl2015deep,ho2020denoising} was first introduced as a probabilistic generative model inspired by nonequilibrium thermodynamics. Subsequently, Song~\etal~\cite{song2021scorebased} reformulated it within the framework of stochastic differential equations (SDEs). In this paper, we propose a general diffusion inversion technique that is applicable to both the probabilistic and SDE-based diffusion formulations. To facilitate understanding, we employ the probabilistic framework of the Denoising Diffusion Probabilistic Model (DDPM)~\cite{ho2020denoising} throughout our presentation.

The DDPM framework~\cite{ho2020denoising} is indeed a Markov chain of length $T$, where the forward process is characterized by a Gaussian transition kernel: 
\begin{equation}
    q(\bm{x}_t|\bm{x}_{t-1}) = \mathcal{N}(\bm{x}_t; \sqrt{1-\beta_t}\bm{x}_{t-1},\beta_t\bm{I}),  \label{eq:ddpm_forward}
\end{equation}
where $\beta_t$ is a pre-defined hyper-parameter controlling variance schedule. Notably, this transition kernel allows the derivation of the marginal distribution $q(\bm{x}_t|\bm{x}_0)$, \ie,
\begin{equation}
    q(\bm{x}_t|\bm{x}_0) = \mathcal{N}(\bm{x}_t; \sqrt{\bar{\alpha}_t}\bm{x}_0, (1-\bar{\alpha}_t)\bm{I}), \label{eq:ddpm_forward_marginal}
\end{equation}
where $\bar{\alpha}_t=\prod_{s=1}^t\alpha_s$, $\alpha_s=1-\beta_s$. The reverse process aims to generate a high-quality image from an initial random noise map $\bm{x}_T \sim \mathcal{N}(\bm{0},\bm{I})$, which can be expressed as:
\begin{equation}  
    \bm{x}_{t-1} = g_{\theta}(\bm{x}_t, t) + \sigma_t \bm{z}_{t-1}, ~ t=T, \cdots,1, \label{eq:ddpm_reverse}
\end{equation}
where 
\begin{equation}  
    g_{\theta}(\bm{x}_t, t) = \frac{1}{\sqrt{\alpha_t}} \left(\bm{x}_t-\frac{1-\alpha_t}{\sqrt{1-\bar{\alpha}_t}}\bm{\epsilon}_{\theta}(\bm{x}_t, t)\right), \label{eq:ddpm_reverse_supp}
\end{equation}
$\bm{\epsilon}_{\theta}(\bm{x}_t, t)$ is a pre-trained denoising network parameterized by $\theta$. The noise term $\bm{z}_t$ satisfies $\bm{z}_0 = \bm{0}$ and $\bm{z}_{t} \sim \mathcal{N}(\bm{0},\bm{I})$ for $t=1,\cdots,T-1$.

Equation~\eqref{eq:ddpm_reverse} indicates that the synthesized image $\bm{x}_0$ is fully determined by the set of noise maps $\mathcal{M}=\bm{x}_T \cup \{\bm{z}_t\}_{t=1}^{T-1}$. In the context of SR, our goal is to generate an HR image $\bm{x}_0$ conditioned on an LR image $\bm{y}_0$. To this end, we propose diffusion inversion to find an optimal set of noise maps $\mathcal{M}^{*}$ that reconstruct the target HR image $\bm{x}_0$ via the reverse process of Eq.~\eqref{eq:ddpm_reverse}. In the following sections, we detail how to achieve this goal by training a noise predictor.

\subsection{Diffusion Inversion}
To achieve diffusion inversion, we introduce a noise prediction network with parameter $w$, denoted as $f_w$, which takes the LR image $\bm{y}_0$ and the timestep $t$ as input and outputs the desired noise maps $\mathcal{M}^{*}$. Unlike the strategy~\cite{chihaoui2024blind} of directly optimizing $\mathcal{M}^{*}$ for each testing image, we train such a noise predictor to enable fast sampling during inference, thereby significantly improving the inference efficiency. To ensure the output of $f_w$ conforms to Gaussian distribution, we adopt the reparameterization trick of VAE~\cite{kingma2014auto}, which predicts the mean and variance parameters of Gaussian distribution rather than directly estimating the noise map.

\subsubsection{Problem Simplification}
Training this noise predictor is inherently challenging. The noise map set $\mathcal{M}$ consists of $T$ noise maps (typically $T=1000$ in most current diffusion models), corresponding to each step of the diffusion process. Naturally, it is non-trivial to simultaneously estimate such a large number of noise maps using a single, compact network. What's worse, the iterative sampling paradigm of diffusion models can gradually accumulate prediction errors, which may adversely affect the final SR performance.

To address these challenges, we design a Partial Noise Prediction (PnP) strategy. Specifically, let's consider diffusion inversion in the context of SR, where the observed LR image $\bm{y}_0$ only slightly deviates from the target HR image $\bm{x}_0$ in most cases, primarily in high-frequency components. This observation inspires us to initiate the sampling process from an intermediate timestep $N$ ($N<T$), effectively reducing the number of noise maps in $\mathcal{M}$ from $T$ to $N$, \ie, $\mathcal{M}=\{\bm{z}_t\}_{t=1}^N$. Furthermore, given the high-fidelity requirements of SR, we constrain $\bm{x}_N$ to have a relatively high SNR, implying mild noise corruption. This constraint encourages the selection of a smaller $N$, and in practice, we set $N \leq 250$, corresponding to an SNR threshold of $1.44$ in the widely used Stable Diffusion~\cite{rombach2022high}.

In addition, we further compress the set of the noise maps $\mathcal{M}=\{\bm{z}_t\}_{t=1}^N$ by integrating existing diffusion acceleration algorithms~\cite{song2021denoising,karras2022elucidating}. The common idea of these algorithms is to skip certain steps during inference, which are selected based on specific rules~\cite{lin2024common}, \eg, ``linspace'' and ``trailing''. Combining with this skipping strategy, the noise map set is simplified as follows:
\begin{equation}
    \mathcal{M} = \{\bm{z}_{\kappa_i}\}_{i=1}^M, \label{eq:noise_map_set}
\end{equation}
where $\{\kappa_1, \cdots, \kappa_M\} \subseteq \{1, \cdots, N\}$. In practice, we set $M\leq 5$, thus largely reducing the prediction burden on the noise predictor and improving the sampling efficiency.

\subsubsection{Inversion Trajectory}
Given the set of noise maps $\mathcal{M} = \{\bm{z}_{\kappa_i}\}_{i=1}^M$ and the noise prediction network $f_w$, our goal is to restore the HR image $\bm{x}_0$ from a given LR observation $\bm{y}_0$, following an inversion trajectory defined by:
\begin{equation}  
    \bm{x}_{\kappa_{i-1}} = g_{\theta}(\bm{x}_{\kappa_i}, \kappa_i) + \sigma_{\kappa_i} f_w(\bm{y}_0,\kappa_{i-1}), \label{eq:ddpm_reverse_noise_predict}
\end{equation}
where $\kappa_0=0$, and $g_{\theta}(\cdot,\cdot)$ is defined in Eq.~\eqref{eq:ddpm_reverse_supp}. The key to initiating this inversion trajectory is constructing the starting state $\bm{x}_{\kappa_M}$ from the LR image $\bm{y}_0$.

The marginal distribution $q(\bm{x}_{\kappa_M}|\bm{x}_0)$, as defined in Eq.~\eqref{eq:ddpm_forward_marginal}, suggests to achieve $\bm{x}_{\kappa_M}$ as follows:
\begin{equation}
    \bm{x}_{\kappa_M} = \sqrt{\bar{\alpha}_{\kappa_M}}\bm{x}_0 + \sqrt{1-\bar{\alpha}_{\kappa_M}} \bm{\xi}, ~\bm{\xi} \sim \mathcal{N}(\bm{0}, \bm{I}). \label{eq:marginal_from_x0}
\end{equation}
In the context of SR, since the HR image $\bm{x}_0$ is not accessible during testing, we thus construct an analogous formulation for $\bm{x}_{\kappa_M}$ directly from the LR image $\bm{y}_0$ using the noise predictor $f_w(\cdot)$, namely
\begin{equation}
    \bm{x}_{\kappa_M} = \sqrt{\bar{\alpha}_{\kappa_M}}\bm{y}_0 + \sqrt{1-\bar{\alpha}_{\kappa_M}} f_w(\bm{y}_0, \kappa_M). \label{eq:marginal_from_y0_M}
\end{equation}
This design is inspired by the observation that the LR image $\bm{y}_0$ and the HR image $\bm{x}_0$ become increasingly indistinguishable when perturbed by Gaussian noise with an appropriate magnitude. Therefore, we aim to seek an optimal noise map $f_w(\bm{y}_0, \kappa_M)$ to perturb $\bm{y}_0$ in such a way that the pre-trained diffusion model can generate the corresponding $\bm{x}_0$ from $\bm{x}_{\kappa_M}$ that is defined in Eq.~\eqref{eq:marginal_from_y0_M}.

To summarize, we establish an inversion trajectory by combining Eqs.~\eqref{eq:ddpm_reverse_noise_predict} and \eqref{eq:marginal_from_y0_M}, which can be used to solve the SR problem via iterative generation along this trajectory. 

\subsubsection{Model Training}\label{subsubsec:model-training}
Given a pre-trained large-scale diffusion model $\bm{\epsilon}_{\theta}(\cdot)$, an estimation of the HR image $\bm{x}_0$ can be obtained from $\bm{x}_{\kappa_i}$ by taking a reverse diffusion step: 
\begin{equation}
    \hat{\bm{x}}_{0\leftarrow\kappa_i} = \frac{1}{\sqrt{\bar{\alpha}_{\kappa_i}}}\left[\bm{x}_{\kappa_i} - \sqrt{1-\bar{\alpha}_{\kappa_i}}\bm{\epsilon}_{\theta}(\bm{x}_{\kappa_i}, \kappa_i)\right], \label{eq:x0_prediction_from_reverse}
\end{equation}
where $\bm{x}_{\kappa_i}$ is defined by Eq.~\eqref{eq:marginal_from_y0_M} for $i=M$ and Eq.~\eqref{eq:ddpm_reverse_noise_predict} for $i<M$. It is thus possible to train the noise predictor $f_w(\cdot)$ by minimizing the distance between $\hat{\bm{x}}_{0\leftarrow\kappa_i}$ and $\bm{x}_0$.   

\begin{figure}
    \centering
    \includegraphics[width=\linewidth]{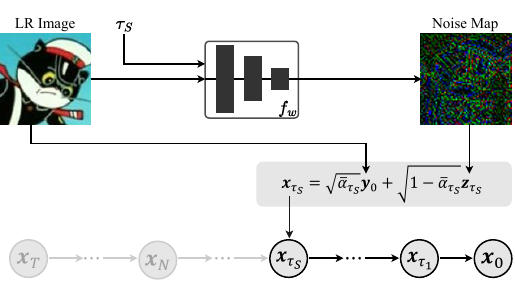}
    \caption{Inference flow of our proposed method, wherein $\{\tau_i\}_{i=1}^S$ denotes the inversion timesteps. Note that the predicted noise map $\bm{z}_{\tau_S}$ exhibits an obvious correlation with the LR image, indicating the non-zero mean property of its statistical distribution.}
    \label{fig:framework}
\end{figure}

However, directly training with this objective is computationally impractical. Specifically, as shown in Eq.~\eqref{eq:ddpm_reverse_noise_predict}, calculating $\bm{x}_{\kappa_i}$ ($i<M$) necessitates recurrent application of the large-scale diffusion model $\bm{\epsilon}_{\theta}$, which leads to prohibitive GPU memory usage. To circumvent this, we adopt an alternative version for $\bm{x}_{\kappa_i}$ based on the marginal distribution in Eq.~\eqref{eq:ddpm_forward_marginal}, \ie,
\begin{equation}
    \bm{x}_{\kappa_i} = \sqrt{\bar{\alpha}_{\kappa_i}}\bm{x}_0 + \sqrt{1-\bar{\alpha}_{\kappa_i}} f_w(\bm{y}_0, \kappa_i), ~ i<M. \label{eq:marginal_from_x0_i}
\end{equation}
This modification also aligns better with the training process of the employed diffusion model, allowing for more effective leveraging of the prior knowledge embedded in it. We now detail the training procedure step by step: 

\vspace{1mm}\noindent\textbf{Gaussian Constraint}. The pre-trained diffusion model is a powerful denoiser tailored for Gaussian noise with zero mean and varying variances. Hence, it is reasonable to enforce the predicted noise map by $f_w$ to obey a Gaussian distribution. For the initial state $\bm{x}_{\kappa_M}$, it is observed that the predicted noise map $f_w(\bm{y}_0,\kappa_M)$ exhibits a mean shift, which is evident when comparing Eqs.~\eqref{eq:marginal_from_x0} and \eqref{eq:marginal_from_y0_M}, due to the substitution of $\bm{y}_0$ for $\bm{x}_0$. Moreover, the visualization presented in Figs.~\ref{fig:framework} and \ref{fig:noise-pred-supp} further validates this observation, illustrating that the predicted noise map is clearly correlated with the LR image. Therefore, we do not consider the Gaussian constraint for $\bm{x}_{\kappa_M}$. 

Conversely, for the intermediate state $\bm{x}_{\kappa_i}$ as defined in Eq.~\eqref{eq:marginal_from_x0_i}, the predicted noise map $f_w(\bm{y}_{\kappa_i},\kappa_i)$ should be enforced to follow $\mathcal{N}(\bm{0},\bm{I})$. 
This naturally raises an interesting question: Is it necessary to predict the noise map using  $f_w$ instead of random sampling?

First, the proposed PnP strategy requires the timestep ${\kappa_i}$ to satisfy a high SNR constraint, indicating that $\bm{x}_{\kappa_i}$ is corrupted by only mild Gaussian noise. Second, the pre-trained large-scale diffusion model, specifically designed for Gaussian denoising, performs robustly, especially for timesteps $\kappa_i$ with low noise levels. Thus, introducing an extra noise predictor model $f_w$ for the intermediate timesteps, even conditioned on the additional LR observation, does not yield significant performance gains. This is also empirically illustrated by an ablation study provided in supp. Third, predicting the noise map both for the initial and intermediate state would increase the prediction burden on $ f_w$ and make the training process more challenging. Considering these reasons together, we discard the noise prediction for the intermediate states, which further reduces the noise map set to $\mathcal{M}=\{\bm{z}_{\kappa_M}\}$, leading to a more elegant diffusion inversion technique, as detailed in Alg.~\ref{alg:inference}. 

\vspace{1mm}\noindent\textbf{Arbitrary-step Inversion}. As analyzed above, noise map prediction is only required for the starting state, as defined in Eq.\eqref{eq:marginal_from_y0_M}, to initialize the reverse sampling process. To further enhance the flexibility of the sampling process, the noise predictor is trained to estimate the noise maps for multiple pre-selected steps via time embedding. Once trained, the starting timestep can be freely chosen during inference, which results in a fidelity-realism trade-off, as analyzed in Sec.~\ref{subsec:analysis-exp}. Note that the total number of sampling steps is determined by both the selected starting timestep and the skipping stride of the accelerated sampling algorithm for diffusion models. For clarity, a detailed inference procedure is provided in Fig.~\ref{fig:framework} and Alg.~\ref{alg:inference}. Given practical efficiency considerations, we focused on the number of sampling steps ranging from one to five in this study.

\begin{algorithm}[t]
    \caption{Inference}
    \label{alg:inference}
    \begin{algorithmic}[1]
        \REQUIRE LR image $\bm{y}_0$, noise predictor $f_w$, pre-trained diffusion model $\bm{\epsilon}_{\theta}$, inversion timesteps $\{\tau_i\}_{i=1}^S\subset \{\kappa_i\}_{i=1}^M$ 
        \STATE $\bm{x}_{\tau_S} = \sqrt{\bar{\alpha}_{\tau_S}}\bm{y}_0 + \sqrt{1-\bar{\alpha}_{\tau_S}} f_w(\bm{y}_0, \tau_S)$ 
        \FOR{$i=S,\cdots,1$}
        \STATE $\bm{z}_{\tau_i} \sim \mathcal{N}(\bm{0},\bm{I})$ if $\tau_i>1$ else $\bm{z}_{\tau_i}=\bm{0}$
        \STATE $\bm{x}_{\tau_{i-1}} = g_{\theta}(\bm{x}_{\tau_i}, \tau_i) + \sigma_{\tau_i} \bm{z}_{\tau_i}$, where $g_{\theta}$ is defined in Eq.~\eqref{eq:ddpm_reverse_supp}
        \ENDFOR
        \RETURN $\bm{x}_0$
    \end{algorithmic}
\end{algorithm}
\vspace{1mm}\noindent\textbf{Loss Function}. To train the noise predictor, we adopt an $L_2$ loss $\mathcal{L}_2$, a LPIPS~\cite{zhang2018unreasonable} loss $\mathcal{L}_l$, and a GAN~\cite{goodfellow2014generative} loss $\mathcal{L}_g$ following recent SR approaches~\cite{wang2021real,chan2022glean}. Let's denote the set of pre-selected starting timesteps for training as $\mathcal{S}\subseteq \{\kappa_1,\cdots,\kappa_M\}$, the overall loss function is defined as:
\begin{small}
 \begin{equation}
    \sum_{t\in \mathcal{S}} \mathcal{L}_2(\hat{\bm{x}}_{0\leftarrow t}, \bm{x}_0) + \lambda_l  \mathcal{L}_l(\hat{\bm{x}}_{0\leftarrow t}, \bm{x}_0) + \lambda_g\mathcal{L}_g(\hat{\bm{x}}_{0\leftarrow t}, \bm{x}_0), \label{eq:loss_total}
\end{equation}   
\end{small}
\hspace{-2mm}where $\hat{\bm{x}}_{0\leftarrow t}$ is defined in Eq.~\eqref{eq:x0_prediction_from_reverse}, $\lambda_l$ and $\lambda_g$ are hyper-parameters. 
The adversarial loss $\mathcal{L}_g$ is implemented using a hinge loss, with a discriminator architecture based on diffusion UNet~\cite{ho2020denoising} enhanced with a multi-in, multi-out strategy~\cite{yin2024improved}. For the base model $\bm{\epsilon}_{\theta}$, we use SD-Turbo~\cite{sauer2024adversarial}, which operates in the latent space of VQGAN~\cite{esser2021taming}. We thus compute the whole loss in the latent space, significantly reducing the required GPU memory. To facilitate training, the LPIPS loss is also fine-tuned in the latent space. 

\begin{table*}[t]
    \centering
    \caption{Quantitative results of \textit{InvSR} with various numbers of sampling steps ranging from one to five on the \textit{ImageNet-Test} dataset.}
    \label{tab:ablation_steps}
    \vspace{-3mm}
    \small
    \begin{tabular}{@{}C{1.4cm}@{}| @{}C{3.6cm}@{}|@{}C{1.7cm}@{} @{}C{1.7cm}@{} @{}C{1.7cm}@{} 
                                    @{}C{1.7cm}@{} @{}C{1.5cm}@{} @{}C{2.0cm}@{} @{}C{2.0cm}@{} }
        \Xhline{0.8pt}
        \multirow{2}*{\makecell{\#Steps}} & \multirow{2}*{\makecell{Index of the sampled\\ timesteps}}  & \multicolumn{7}{c}{Metrics} \\
         \Xcline{3-9}{0.4pt}
             &  & PSNR$\uparrow$   & SSIM$\uparrow$   & LPIPS$\downarrow$   & NIQE$\downarrow$   & PI$\downarrow$   & CLIPIQA$\uparrow$ & MUSIQ$\uparrow$  \\
        \Xhline{0.4pt}
        5  & \{250, 200, 150, 100, 50\}  & 22.70  & 0.6412   & 0.2844   & 4.8757  & 3.4744  & 0.6733 & 69.8427 \\
        \hline\hline
         \multirow{3}*{3}  & \{250, 150, 50\}   & 22.92  & 0.6478   & 0.2762   & 4.7980  & 3.4002  & 0.6823 & 70.4688 \\
         & \{200, 100, 50\}   & 23.41      & 0.6609        & 0.2648        & 4.5089       & 3.2074  & 0.6851 & 70.7024 \\
         & \{150, 100, 50\}   & 23.84  & 0.6713   & 0.2575   & 4.2719  & 3.0527  & 0.6823 & 70.4569 \\
        \hline\hline
        \multirow{4}*{1}  & \{250\}            & 23.84  & 0.6713   & 0.2575   & 4.5287  & 3.1748  & 0.7132 & 72.5773 \\
          & \{200\}            & 24.14  & 0.6789   & 0.2517   & 4.3815  & 3.0866  & 0.7093 & 72.2909 \\
          & \{150\}            & 24.42      & 0.6851        & 0.2469        & 4.2194       & 2.9979      & 0.7019      & 71.7100  \\
          & \{100\}            & 24.66  & 0.6891   & 0.2450   & 4.0606  & 2.8951  & 0.6912 & 70.8251  \\
        \Xhline{0.8pt}
    \end{tabular}
\end{table*}
\section{Experiments}
In this section, we first provide an analysis of the proposed method and then conduct extensive experiments to evaluate its performance on one synthetic and two real-world datasets. Our investigation focuses mainly on the $\times4$ SR task following previous works~\cite{zhang2021designing,wang2021real}. To ease the presentation, we refer to our method as \textit{InvSR}, standing for Diffusion \textbf{\textit{Inv}}ersion-based \textbf{\textit{S}}uper-\textbf{\textit{R}}esolution. 

\subsection{Experimental Setup}
\textbf{Training Details}. Following the setup of recent works~\cite{wu2024seesr,wu2024one}, we trained the noise predictor on the LSDIR~\cite{li2023lsdir} dataset and a subset of 20k face images from the FFHQ~\cite{karras2019style} dataset. At each iteration, we randomly cropped an image patch with a resolution of $512\times 512$ from the source image and synthesized the LR image using the pipeline of RealESRGAN~\cite{wang2021real}. The text prompt was fixed as a general description\footnote{``Cinematic, high-contrast, photo-realistic, 8k, ultra HD, meticulous detailing, hyper sharpness, perfect without deformations.''} in both the training and testing phases. To optimize the network parameters, we adopted the Adam~\cite{kingma2015adam} algorithm with default settings of PyTorch~\cite{paszke2019pytorch}. The training process takes over 100k iterations with a batch size of 64 and a fixed learning rate of $5\text{e}^{-5}$. The hyper-parameters $\lambda_l$ and $\lambda_g$ in the loss function were set to 2.0 and 0.1, respectively. The architecture of the noise predictor was based on the encoder of VQGAN~\cite{esser2021taming}, containing two down-sampling blocks, each equipped with a self-attention layer~\cite{vaswani2017attention}. 

In the training stage, we randomly select a starting timestep from $\mathcal{S}=\{250,200,150,100\}$ to train the noise predictor at each iteration. During inference, five inversion steps, \ie, $\mathcal{M}=\{250,200,150,100,50\}$, are used throughout our experiments. 

\vspace{1mm}\noindent\textbf{Testing Datasets}. To evaluate the performance of \textit{InvSR}, we constructed a synthetic dataset named \textit{ImageNet-Test}, comprising 3,000 images from the validation set of ImageNet~\cite{deng2009imagenet}. The LR and HR images, with resolutions of $128\times 128$ and $512 \times 512$, respectively, were synthesized using the degradation settings of ResShift~\cite{yue2024efficient}. Notably, we selected the HR images from ImageNet rather than the commonly used datasets in SR, such as Set5~\cite{bevilacqua2012low}, Set14~\cite{zeyde2012single}, and Urban100~\cite{huang2015single}, mainly because these datasets only contain very few source images, which fails to thoroughly assess various methods under
complicated degradation types.

We further conducted experiments on two real-world datasets to validate the effectiveness of \textit{InvSR}.
The first dataset is \textit{RealSR}~\cite{cai2019toward}, which contains 100 real images captured by Canon 5D3 and Nikon D810 cameras. The second dataset, \textit{RealSet80}~\cite{yue2024efficient}, comprises 80 LR images widely used in existing literature~\cite{martin2001database,matsui2017sketch,Ignatov2017DSLR,wang2021real,lin2023diffbir,yue2023resshift}. 

\vspace{1mm}\noindent\textbf{Compared Methods}. We evaluate the effectiveness of \textit{InvSR} in comparison to nine recent methods, including two GAN-based methods, namely BSRGAN~\cite{zhang2021designing} and RealESRGAN~\cite{wang2021real}, as well as seven diffusion-based methods, including LDM~\cite{rombach2022high}, StableSR~\cite{wang2024exploiting}, DiffBIR~\cite{lin2023diffbir}, SeeSR~\cite{wu2024seesr}, ResShift~\cite{yue2023resshift,yue2024efficient}, SinSR~\cite{wang2024sinsr}, and OSEDiff~\cite{wu2024one}. For LDM, StableSR, DiffBIR, and SeeSR, we all use 50 sampling steps for fair comparison. In the case of ResShift, SinSR, and OSEDiff, we adhere to the number of sampling steps suggested by their official guidelines.

\vspace{1mm}\noindent\textbf{Metrics}. The performance of various methods was assessed across seven metrics, including three reference metrics, namely PSNR, SSIM~\cite{zhou2004image}, LPIPS~\cite{zhang2018unreasonable}, as well as four non-reference metrics, namely NIQE~\cite{mittal2012making}, PI~\cite{blau20182018}, MUSIQ~\cite{ke2021musiq}, and CLIPIQA~\cite{wang2023exploring}. For evaluations on the datasets of \textit{ImageNet-Test} and \textit{RealSR}, all seven metrics were adopted to ensure a holistic assessment. For the dataset of \textit{RealSet80}, however, only non-reference metrics were employed since the ground truth images are not accessible. Notably, PSNR and SSIM are calculated in the luminance (Y) channel of YCbCr space, while other metrics are directly computed in the standard RGB (sRGB) space. 

\begin{table*}[t]
    \centering
    \caption{Quantitative comparisons of different methods on \textit{ImageNet-Test} and \textit{RealSR}. The number of sampling steps is marked in the format of ``Method name-Steps'' for diffusion-based methods. The best and second-best results are highlighted in \textbf{bold} and \underline{underlined}.}
    \label{tab:metric_imagenet}
    \vspace{-3mm}
    \small
    \begin{tabular}{@{}C{2.2cm}@{}| @{}C{2.8cm}@{}|@{}C{1.4cm}@{} @{}C{1.4cm}@{} @{}C{1.4cm}@{} 
                                    @{}C{1.4cm}@{} @{}C{1.4cm}@{} @{}C{1.8cm}@{} @{}C{1.8cm}@{} @{}C{1.8cm}@{} }
        \Xhline{0.8pt}
        \multirow{2}*{Datasets} &\multirow{2}*{Methods}  & \multicolumn{8}{c}{Metrics} \\
         \Xcline{3-10}{0.4pt}
              & & PSNR$\uparrow$   & SSIM$\uparrow$   & LPIPS$\downarrow$   & NIQE$\downarrow$   & PI$\downarrow$   & CLIPIQA$\uparrow$ & MUSIQ$\uparrow$ & \#Params (M)  \\
        \Xhline{0.4pt}
       \multirow{10}*{\textit{ImageNet-Test}} &BSRGAN~\cite{zhang2021designing}   & 27.05  & 0.7453   & 0.2437   & 4.5345  & 3.7111  & 0.5703  & 67.7195 & 16.70 \\ 
       & RealESRGAN~\cite{wang2021real}   &  26.62 & \underline{0.7523} & 0.2303 & 4.4909 & 3.7234 & 0.5090 & 64.8186 & 16.70 \\
       \Xcline{2-10}{0.4pt}
       & LDM-50~\cite{rombach2022high}   & \underline{27.19}  & 0.7285 & 0.2286 & 5.2411 & 4.2554 & 0.5554 & 62.8776 & 113.60 \\
       & StableSR-50~\cite{wang2024exploiting} & 24.77 & 0.6908 & 0.2591 & 4.5120 & \underline{3.1473} & \underline{0.7067} & \underline{71.2811}  & 152.70\\
       & DiffBIR-50~\cite{lin2023diffbir}  & 25.72  & 0.6695 & 0.2795 & 4.5875 & 3.2260 & 0.6900 & 69.7089 & 385.43  \\
       & SeeSR-50~\cite{wu2024seesr}     &26.69     & 0.7422 & \underline{0.2187} & \underline{4.3825} & 3.4742 & 0.5868 & 71.2412 & 751.50  \\
       \Xcline{2-10}{0.4pt}
       & ResShift-4~\cite{yue2024efficient} & \textbf{27.33} & \textbf{0.7530} & \textbf{0.1998} & 5.8700 & 4.3643 & 0.6147 & 65.5860 & 118.59 \\
       \Xcline{2-10}{0.4pt}
       & SinSR-1~\cite{wang2024sinsr}         & 26.98 & 0.7304 & 0.2209 & 5.2623 & 3.8189 & 0.6618 & 67.7593 & 118.59 \\
       & OSEDiff-1~\cite{wu2024one}           & 23.95 &0.6756  & 0.2624 & 4.7157 & 3.3775 & 0.6818 & 70.3928  & 8.50\\
       & \textit{InvSR}-1 (Ours)       & 24.14 & 0.6789 & 0.2517 & \textbf{4.3815} & \textbf{3.0866} & \textbf{0.7093} & \textbf{72.2900} & 33.84 \\
       \hline \hline
        \multirow{10}*{\textit{RealSR}} &BSRGAN~\cite{zhang2021designing}   & \underline{26.51}  & \underline{0.7746}   & \underline{0.2685}   & 4.6501  & 4.4644  & 0.5439  & 63.5869 & 16.70\\ 
       & RealESRGAN~\cite{wang2021real}   &  25.85 & 0.7734 & 0.2728 & 4.6766 & 4.4881 & 0.4898 & 59.6803 & 16.70 \\
       \Xcline{2-10}{0.4pt}
       & LDM-50~\cite{rombach2022high}   & \textbf{26.75}  & 0.7711 & 0.2945 & 4.8712 & 5.0025 & 0.4907 & 54.3910 & 113.60 \\
       & StableSR-50~\cite{wang2024exploiting} & 26.27 & \textbf{0.7755} & \textbf{0.2671} & 5.1745 & 4.8209 & 0.5209 & 60.1758 & 152.70 \\
       & DiffBIR-50~\cite{lin2023diffbir}  & 24.83  & 0.6642 & 0.3864 & \textbf{3.7366} & \textbf{3.3661} & 0.6857 & 65.3934  & 385.43\\
       & SeeSR-50~\cite{wu2024seesr}     &26.20     & 0.7555 & 0.2806 & 4.5358 & 4.1464 & 0.6824 & \underline{66.3757} & 751.50 \\
       \Xcline{2-10}{0.4pt}
       & ResShift-4~\cite{yue2024efficient} & 25.77 & 0.7453 & 0.3395 & 6.9113 & 5.4013 & 0.5994 & 57.5536 & 118.59 \\
       \Xcline{2-10}{0.4pt} 
       & SinSR-1~\cite{wang2024sinsr}       & 26.02 & 0.7097 & 0.3993 & 6.2547 & 4.7183 & 0.6634 & 59.2981  & 118.59 \\
       & OSEDiff-1~\cite{wu2024one}           & 23.89 &0.7030  & 0.3288 & 5.3310 & 4.3584 & \textbf{0.7008} & 65.4806 & 8.50  \\
       & \textit{InvSR}-1 (Ours)       & 24.50 & 0.7262 & 0.2872 & \underline{4.2189} & \underline{3.7779} & \underline{0.6918} & \textbf{67.4586} & 33.84 \\
       \Xhline{0.8pt} 
    \end{tabular}
\end{table*}
\subsection{Model Analysis} \label{subsec:analysis-exp}
\noindent\textbf{Arbitrary-steps Sampling}. Recent efficient diffusion-based SR approaches, such as ResShift~\cite{yue2024efficient}, SinSR~\cite{wang2024sinsr}, and OSEDiff~\cite{wu2024one}, 
constrain the sampling process to a predefined number of steps, consistent with their training configuration. In contrast, the proposed \textit{InvSR} supports sampling with an arbitrary number of steps, significantly enhancing flexibility and adaptability to varying degradation types, as demonstrated in Fig.~\ref{fig:teaser} and Fig.~\ref{fig:multi-step-supp}. 
\begin{figure}[t]
    \centering
    \includegraphics[width=\linewidth]{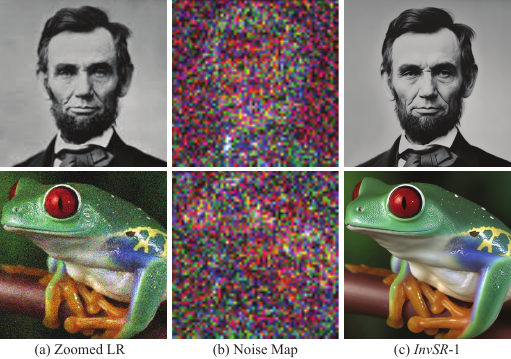}
    \vspace{-6mm}
    \caption{From left to right: (a) zoomed LR image, (b) predicted noise map by our method for the initial timestep, (c) super-resolved results by our method with a single sampling step.}
    \label{fig:noise-pred-supp}
\end{figure}

We further provide a comprehensive comparison of \mbox{\textit{InvSR}} with one, three, and five sampling steps, as summarized in Table.~\ref{tab:ablation_steps}. Three key observations can be obtained from these results: i) With a fixed number of sampling steps, \eg, one or three, varying the starting timestep enables a trade-off between fidelity (measured by reference metrics) and realism (measured by non-reference metrics). Specifically, using larger starting timesteps favors improved realism at the expense of fidelity. ii) As expected, reference metrics deteriorate with increased sampling steps due to the introduction of additional randomness. iii) Interestingly, non-reference metrics also exhibit a decline when using more sampling steps. This is mainly because most testing images contain some noise, which can lead to undesired artifacts if multiple sampling steps are used, thus degrading the overall image quality. However, using more sampling steps can effectively recover intricate fine-grained structures in cases involving substantial blur, as evidenced by the first examples in Fig.~\ref{fig:teaser} and Fig.~\ref{fig:multi-step-supp}.

\vspace{1mm}\noindent\textbf{Initial Noise Prediction} Figure~\ref{fig:noise-pred-supp} presents the noise map predicted by our method for the initial timestep, exhibiting a strong correlation with image content. This visualization aligns well with the theoretical analysis in Sec~\ref{subsubsec:model-training}, empirically validating that our noise predictor can effectively find an LR-dependent noise map to facilitate the SR task.

\subsection{Comparison to State of the Arts}
Considering that recent studies~\cite{wang2024sinsr,wu2024one} mainly focus on developing one-step diffusion-based methods, we thus evaluate \textit{InvSR} against these methods under a one-step configuration to ensure a fair comparison.

\begin{table}[t]
    \centering
    \caption{Quantitative comparisons of various methods on \textit{RealSet80}. The number of sampling steps is marked in the format of ``Method name-Steps'' for diffusion-based methods. The best and second-best results are highlighted in \textbf{bold} and \underline{underlined}. }
    \vspace{-3mm}
    \label{tab:metric_realset80}
    \small
    \begin{tabular}{@{}C{2.6cm}@{}| @{}C{1.3cm}@{} @{}C{1.2cm}@{} @{}C{1.6cm}@{} @{}C{1.6cm}@{} }
        \Xhline{0.8pt}
        \multirow{2}*{Methods}  & \multicolumn{4}{c}{Metrics} \\
         \Xcline{2-5}{0.4pt}
              & NIQE$\downarrow$   & PI$\downarrow$   & CLIPIQA$\uparrow$ & MUSIQ$\uparrow$  \\
        \Xhline{0.4pt}
       BSRGAN~\cite{zhang2021designing}   & 4.4408  & 4.0276  & 0.6263  & 66.6288\\ 
       RealESRGAN~\cite{wang2021real}     & 4.1568  & 3.8852  & 0.6189  & 64.4957 \\
       \Xhline{0.4pt}
       LDM-50~\cite{rombach2022high}      & 4.3248  & 4.2545  & 0.5511  & 55.8246 \\
       StableSR-50~\cite{wang2024exploiting} & 4.5593 & 4.0977 & 0.6214 & 62.7613 \\
       DiffBIR-50~\cite{lin2023diffbir}   & \textbf{3.8630}  & \textbf{3.2117}  & \textbf{0.7404}  & 67.9806 \\
       SeeSR-50~\cite{wu2024seesr}        & 4.3678  & 3.7429  & 0.7114  & \underline{69.7658} \\
       \Xhline{0.4pt}
       ResShift-4~\cite{yue2024efficient} & 5.9866  & 4.8318  & 0.6515  & 61.7967 \\
       \Xhline{0.4pt}
       SinSR-1~\cite{wang2024sinsr}       & 5.6243  & 4.2830  & 0.7228  & 64.0573 \\
       OSEDiff-1~\cite{wu2024one}         & 4.3457  & 3.8219  & 0.7093  & 68.8202 \\
       \textit{InvSR}-1 (Ours)            & \underline{4.0284}    & \underline{3.4666} & \underline{0.7291} & \textbf{69.8055} \\
       \Xhline{0.8pt} 
    \end{tabular}
    \vspace{-2mm}
\end{table}
\begin{figure*}[t]
    \centering
    \includegraphics[width=1.00\linewidth]{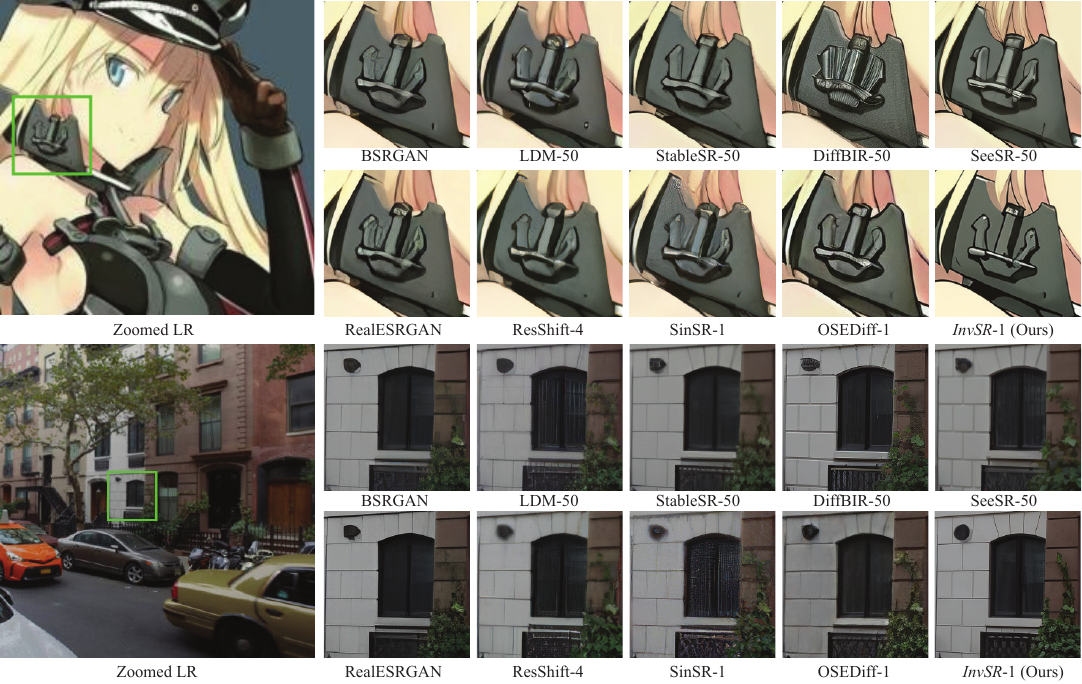}
    \vspace{-6mm}
    \caption{Visual results of different methods on two typical real-world examples from \textit{RealSet80} dataset. For clear comparisons, the number of sampling steps is annotated in the format ``Method name-Steps'' for diffusion-based approaches. \textit{(Zoom-in for best view)}}
    \label{fig:realset80}
\end{figure*}
\vspace{1mm}\noindent\textbf{Synthetic Dataset}. Table~\ref{tab:metric_imagenet} reports a comprehensive evaluation of various methods on \textit{ImageNet-Test} dataset, encompassing seven quantitative metrics, with additional qualitative comparisons included in Fig.~\ref{fig:imagenet-supp} of supp. Notably, compared to the recent state-of-the-art (SotA) one-step method OSEDiff~\cite{wu2024one}, \textit{InvSR} demonstrates evident superiority across all the seven metrics. Moreover, even compared to multi-step methods with 50 sampling steps, such as StableSR and DiffBIR, \textit{InvSR} still achieves comparable performance in distortion-oriented metrics, including PSNR and SSIM, while outperforming these methods in perception-oriented metrics, such as LPIPS, NIQE, PI, and MUSIQ. These results indicate that \textit{InvSR} effectively balances both performance and efficiency, advancing the field of diffusion-based SR approaches. Additionally, \textit{InvSR} maintains a moderate model size with about 34 million learnable parameters, further enhancing its practicality for real-world applications.

\vspace{1mm}\noindent\textbf{Real-world Dataset}. To evaluate real-world datasets, we mainly focus on the non-reference metrics. Table~\ref{tab:metric_imagenet} and \ref{tab:metric_realset80} provide a detailed comparison of \textit{InvSR} against recent SotA methods on the datasets of \textit{RealSR} and \textit{RealSet80}, respectively. It can be easily observed that \textit{InvSR} achieves superior performance across most non-reference metrics compared to recent one-step methods under fair comparison and second-best results compared to existing multi-step methods. To further substantiate these conclusions, we present visual comparisons of two real-world examples in Fig.~\ref{fig:realset80}, and more examples can be found in Fig.~\ref{fig:realset80-supp} of supp. In the first example, where the LR image contains evident compression noise, \textit{InvSR} successfully removes these artifacts and generates clear results, while other methods struggle with remaining artifacts. In the second example, which is degraded by noticeable blurriness, \textit{InvSR} produces sharper image structures, such as the tile edges on the wall. These quantitative and qualitative evaluations highlight the great potential of \textit{InvSR} to solve the real-world SR task.

\section{Conclusion}
We proposed \textit{InvSR}, a new SR method based on diffusion inversion. Our method introduces a noise prediction network designed to estimate an optimal noise map, enabling the construction of an intermediate state of a pre-trained diffusion model as the starting sampling point. This design is appealing in two aspects: first, \textit{InvSR} can sufficiently harness the prior knowledge encapsulated in the pre-trained diffusion model, thereby facilitating SR performance. Second, \textit{InvSR} offers a flexible sampling strategy capable of starting from various intermediate states of the diffusion model by incorporating a time-dependent architecture of the noise predictor. This flexibility allows users to freely adjust the sampling steps according to the degradation type or their specific requirements. Even after reducing the sampling steps to just one, \textit{InvSR} still exhibits significant superiority beyond recent one-step diffusion-based methods, suggesting its effectiveness and efficiency.

\noindent\textbf{Acknowledgement.} This study is supported under the RIE2020 Industry Alignment Fund – Industry Collaboration Projects (IAF-ICP) Funding Initiative, as well as cash and in-kind contributions from the industry partner(s). The work was completed when Zongsheng Yue was with S-Lab.

{
    \small
    \bibliographystyle{ieeenat_fullname}
    \bibliography{main}
}

\clearpage
\appendix
\maketitlesupplementary

\noindent This supplemental material mainly contains:
\vspace{4pt}
\begin{itemize}
    \item Sec.~\ref{sec:discusion_steps} discusses the selection of the number of sampling steps.
    \item Performance comparison of \textit{InvSR} with various base diffusion models in Sec.~\ref{subsec:base_model}.
    \item Ablation studies on the intermediate noise prediction model in Sec.~\ref{subsec:intermeidate_noise_predict}.
    \item Ablation studies on the loss function in Sec.~\ref{subsec:loss_ablation_supp}.
    \item Discussions on the efficiency and limitation in Sec.~\ref{subsec:efficiency_limitation}.
    \item Visual comparisons on \textit{ImageNet-Test} dataset in Fig.~\ref{fig:imagenet-supp}.
    \item More visual comparisons on real-world examples in Fig.~\ref{fig:realset80-supp}.
\end{itemize}

\section{Discussion on Sampling Steps} \label{sec:discusion_steps}
The proposed method, named \textit{InvSR}, enables a flexible sampling mechanism that allows an arbitrary number of sampling steps. This naturally raises an interesting question: how do we determine an appropriate number of sampling steps for general image super-resolution (SR) tasks? We answer this question from two aspects.

First, as shown in Tables 2 and 3, and Fig.~3 of the main text, \textit{InvSR} achieves promising results with only a single sampling step, evidently outperforming recent state-of-the-art (SotA) one-step methods. Therefore, we recommend setting the sampling steps to one for most real-world applications, effectively balancing efficiency and performance.

Second, we can also adjust sampling steps according to the type of image degradation. Generally, image degradations can be categorized into two main classes: blurriness and noise. As illustrated in Fig.~1 and Fig.~\ref{fig:multi-step-supp}, multi-step sampling would incorrectly amplify noise, leading to undesirable artifacts for images with heavy noise. In contrast, for images primarily suffering from blurriness, multi-step sampling proves beneficial, as it generates more detailed and realistic image structures. In practice, we could first estimate the noise level using some off-the-shelf degradation estimation models, such as Mou \etal~\cite{mou2022metric}. Based on the estimated noise level, one can determine whether a one-step or multi-step sampling is more appropriate. In cases where multi-step sampling is favored, the number of sampling steps can be freely adjusted to achieve a satisfactory result.

\begin{figure}[t]
    \centering
    \includegraphics[width=\linewidth]{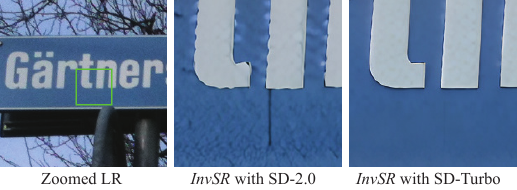}
    \vspace{-6mm}
    \caption{A typical visual comparison of the proposed \textit{InvSR} based on different diffusion models: SD-2.0 and SD-Turbo. Note that these results are achieved with five sampling steps.}
    \label{fig:basemodel-supp}
    \vspace{-2mm}
\end{figure}

\section{Experiments}
\subsection{Base Diffusion Model} \label{subsec:base_model} 
For the pre-trained diffusion models used in \textit{InvSR}, we considered two prevailing variants of Stable Diffusion~\cite{rombach2022high}, namely SD-2.0\footnote{https://huggingface.co/stabilityai/stable-diffusion-2-base} and SD-Turbo\footnote{https://huggingface.co/stabilityai/sd-turbo}. 
Table~\ref{tab:base-model-supp} provides a quantitative comparison of \textit{InvSR} equipped with these two base models. When reducing the sampling steps to one, \textit{InvSR} demonstrated similar performance with both SD-2.0 and SD-Turbo. However, in the multi-step sampling scenarios, the model based on SD-Turbo exhibited more stable performance, particularly in terms of reference metrics. Furthermore, a visual comparison under five sampling steps, as illustrated in Fig.~\ref{fig:basemodel-supp}, reveals that the SD-2.0-based model produced noticeable artifacts, aligning with the quantitative results. We thus employed SD-Turbo as our base model throughout this study.

\begin{table*}[t]
    \centering
    \caption{Quantitative comparisons of the proposed \textit{InvSR} equipped with two different based models, namely SD-2.0 and SD-Turbo, on the dataset of \textit{ImageNet-Test}.}
    \label{tab:base-model-supp}
    \vspace{-3mm}
    \small
    \begin{tabular}{@{}C{1.8cm}@{} | @{}C{1.4cm}@{}| @{}C{3.4cm}@{}|
                                     @{}C{1.5cm}@{} @{}C{1.5cm}@{} @{}C{1.5cm}@{} @{}C{1.5cm}@{} @{}C{1.4cm}@{} @{}C{1.7cm}@{} @{}C{1.7cm}@{} }
        \Xhline{0.8pt}
        \multirow{2}*{Base models} & \multirow{2}*{\makecell{\#Steps}} & \multirow{2}*{\makecell{Index of the sampled\\ timesteps}}  & \multicolumn{7}{c}{Metrics} \\
         \Xcline{4-10}{0.4pt}
            &  &  & PSNR$\uparrow$   & SSIM$\uparrow$   & LPIPS$\downarrow$   & NIQE$\downarrow$   & PI$\downarrow$   & CLIPIQA$\uparrow$ & MUSIQ$\uparrow$  \\
        \Xhline{0.4pt}
        SD-Turbo & \multirow{2}*{5}  & \multirow{2}*{\{250, 200, 150, 100, 50\}}  & 22.70  & 0.6412   & 0.2844   & 4.8757  & 3.4744  & 0.6733 & 69.8427 \\
        SD-2.0 & & & 21.40  & 0.6063   & 0.3274   & 5.1508  & 3.8709  & 0.6467 & 67.6056 \\
        \hline\hline
        SD-Turbo & \multirow{2}*{3}  & \multirow{2}*{\{150, 100, 50\}}  & 23.84  & 0.6713   & 0.2575   & 4.2719  & 3.0527  & 0.6823 & 70.4569 \\
        SD-2.0 & & & 23.13  & 0.6566   & 0.2776  & 4.2449  & 3.1467   & 0.6722 & 69.5178 \\
        \hline\hline
        SD-Turbo & \multirow{2}*{1}  & \multirow{2}*{\{200\}}  & 24.14  & 0.6789   & 0.2517   & 4.3815  & 3.0866  & 0.7093 & 72.2909 \\
        SD-2.0 & & & 23.36  & 0.6637   & 0.2647   & 4.3304  & 3.1545  & 0.6969 & 71.4974 \\
        \Xhline{0.8pt}
    \end{tabular}
\end{table*}
\begin{table*}[t]
    \centering
    \caption{Quantitative comparisons of \textit{InvSR} to the baseline method \textit{InvSR-Int} that combines an additional noise predictor for the intermediate timesteps on the dataset of \textit{ImageNet-Test}.}
    \label{tab:intermediate-noise-pred-supp}
    \vspace{-3mm}
    \small
    \begin{tabular}{@{}C{1.8cm}@{} | @{}C{1.2cm}@{}| @{}C{3.6cm}@{}|
                                     @{}C{1.4cm}@{} @{}C{1.4cm}@{} @{}C{1.5cm}@{} @{}C{1.5cm}@{} @{}C{1.4cm}@{} @{}C{1.8cm}@{} @{}C{1.8cm}@{} }
        \Xhline{0.8pt}
        \multirow{2}*{Methods} & \multirow{2}*{\makecell{\#Steps}} & \multirow{2}*{\makecell{Index of the sampled\\ timesteps}}  & \multicolumn{7}{c}{Metrics} \\
         \Xcline{4-10}{0.4pt}
            &  &  & PSNR$\uparrow$   & SSIM$\uparrow$   & LPIPS$\downarrow$   & NIQE$\downarrow$   & PI$\downarrow$   & CLIPIQA$\uparrow$ & MUSIQ$\uparrow$  \\
        \Xhline{0.4pt}
        \textit{InvSR} & \multirow{2}*{5}  & \multirow{2}*{\{250, 200, 150, 100, 50\}}  & 22.70  & 0.6412   & 0.2844   & 4.8757  & 3.4744  & 0.6733 & 69.8427 \\
        \textit{InvSR-Int} & & & 22.70  & 0.6412   & 0.2844  & 4.8785  & 3.4718  & 0.6734 & 69.8466 \\
        \Xhline{0.8pt}
    \end{tabular}
\end{table*}
\begin{table*}[!t]
    \centering
    \caption{Quantitative ablation studies on the loss function in Eq.~\eqref{eq:loss_total}, wherein the hyper-parameters $\lambda_l$ and $\lambda_g$ control the weight importance of the LPIPS loss and the GAN loss, respectively.}
    \label{tab:loss-ablation-supp}
    \vspace{-3mm}
    \small
    \begin{tabular}{@{}C{1.8cm}@{} | @{}C{2.4cm}@{}| @{}C{2.4cm}@{}|
                                     @{}C{1.4cm}@{} @{}C{1.4cm}@{} @{}C{1.5cm}@{} @{}C{1.5cm}@{} @{}C{1.4cm}@{} @{}C{1.8cm}@{} @{}C{1.8cm}@{} }
        \Xhline{0.8pt}
        \multirow{2}*{Methods} & \multicolumn{2}{c|}{Hyper-parameters}  & \multicolumn{7}{c}{Metrics} \\
         \Xcline{2-10}{0.4pt}
            & $\lambda_l$ (LPIPS loss) & $\lambda_g$ (GAN loss) & PSNR$\uparrow$   & SSIM$\uparrow$   & LPIPS$\downarrow$   & NIQE$\downarrow$   & PI$\downarrow$   & CLIPIQA$\uparrow$ & MUSIQ$\uparrow$  \\
        \Xhline{0.4pt}
        Baseline1 & 0.0  & 0.0  & 26.71  & 0.7365   & 0.2850   & 9.2792  & 6.4147  & 0.6168 & 64.6069 \\
        Baseline2 & 2.0  & 0.0  & 26.24  & 0.7274   & 0.2841   & 8.4367  & 5.7973  & 0.6501 & 66.1726 \\
        Baseline3 & 0.0  & 0.1  & 24.11  & 0.6809   & 0.2599   & 4.4518  & 3.1229  & 0.7078 & 72.5045 \\
        \textit{InvSR-1} & 2.0  & 0.1  & 24.14  & 0.6789   & 0.2517   & 4.3815  & 3.0866  & 0.7093 & 72.2909 \\
        \Xhline{0.8pt}
    \end{tabular}
\end{table*}
\begin{table*}[!t]
    \centering
    \caption{Efficiency comparisons of different methods on the x4 ($128\rightarrow 512$) SR task, where the runtime results are tested on an NVIDIA A100 GPU with 40GB memory. For diffusion-based SR approaches, the number of sampling steps is annotated in the format of ``Method name-Steps''.}
    \label{tab:runtime-supp}
    \vspace{-3mm}
    \small
    \begin{tabular}{@{}C{2.2cm}@{} | @{}C{1.6cm}@{} @{}C{2.3cm}@{} @{}C{1.9cm}@{} @{}C{1.8cm}@{} 
                                     @{}C{1.6cm}@{} @{}C{1.6cm}@{} @{}C{1.5cm}@{} @{}C{1.5cm}@{}
                                     @{}C{1.5cm}@{}}
        \Xhline{0.8pt}
        \multirow{2}*{Metrics}  & \multicolumn{9}{c}{Methods} \\
         \Xcline{2-10}{0.4pt}
         & BSRGAN  & RealESRGAN   & StableSR-50   & DiffBIR-50   & SeeSR-50  & ResShift-4   & SinSR-1   & OSEDiff-1 & \textit{InvSR}-1  \\
        \Xhline{0.4pt}
        \#Params (M)  & 16.70  & 16.70    & 152.70  & 385.43  & 751.50  & 118.59 & 118.59 & 8.50 & 33.84 \\
        Runtime (ms)  & 65  & 65   & 3459  & 7937  & 6438  & 319 & 138  & 176 & 117\\
        \Xhline{0.8pt}
    \end{tabular}
\end{table*}
\subsection{Intermediate Noise Prediction}\label{subsec:intermeidate_noise_predict}
In our proposed diffusion inversion framework, we opt to sample the noise maps randomly rather than employing a noise prediction model for intermediate timesteps. This choice is motivated by the high SNR (signal-to-noise ratio) constraint imposed on the inversion timesteps, as elaborated in Sec.~3.2.3 of the main text. To further validate this choice, we introduced an additional baseline, denoted as ``\textit{InvSR-Int}'', which integrates an extra noise predictor specifically trained for intermediate timesteps. Table~\ref{tab:intermediate-noise-pred-supp} reports a detailed comparison between \textit{InvSR} and \textit{InvSR-Int}. It can be observed that the performance differences between these two models are negligible. Therefore, we omit the intermediate noise prediction in \textit{InvSR}, further simplifying the overall framework.

\subsection{Loss Functions} \label{subsec:loss_ablation_supp}
In addition to the commonly used $L_2$ loss, we incorporate LPIPS~\cite{zhang2018unreasonable} loss and GAN~\cite{goodfellow2014generative} loss to train our noise predictor, as formulated in Eq.~\eqref{eq:loss_total}. The hyper-parameters of $\lambda_l$ and $\lambda_g$ are introduced to control the importance of the LPIPS and GAN losses, respectively. Table~\ref{tab:loss-ablation-supp} provides a quantitative comparison of various baseline models under different loss configurations, and Fig.~\ref{fig:loss-ablation} demonstrates a typical visual example. We can observed that Baseline1 trained solely with the $L_2$-based diffusion loss produces over-smooth outputs, which is consistent with its superior PSNR scores. Incorporating the GAN loss enhances the generation of finer image details but may introduce undesirable artifacts. The addition of the LPIPS loss can mitigate these artifacts to a certain extent, striking a balance between perceptual quality and artifact suppression. Therefore, this study employs both LPIPS and GAN losses to achieve optimal performance.  

\subsection{Efficiency and Limitation}\label{subsec:efficiency_limitation}
Table~\ref{tab:runtime-supp} lists an efficiency comparison of various methods on the x4 ($128\rightarrow 512$) SR task. It can be observed that the proposed \textit{InvSR} demonstrates advantages in runtime among one-step diffusion-based approaches. Despite having a larger number of parameters compared to the recent SotA method OSEDiff~\cite{wu2024one}, \textit{InvSR} achieves a 50\% reduction in inference time. This is mainly because OSEDiff relies on an additional image captioning model, whereas \mbox{\textit{InvSR}} does not. However, it is noteworthy that \textit{InvSR} still lags behind GAN-based methods in efficiency due to its reliance on the large-scale Stable Diffusion model. To address the high-efficiency demand in real-world applications, future work will explore model quantization techniques to further accelerate the inference speed.

\begin{figure*}
    \centering
    \includegraphics[width=\linewidth]{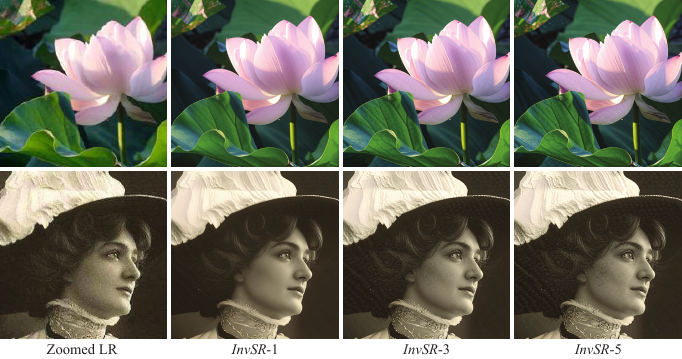}
    \vspace{-6mm}
    \caption{Qualitative comparisons of the proposed \textit{InvSR} with different sampling steps, where the number of sampling steps is annotated in the format ``\textit{InvSR}-Steps''. In the first example, mainly degraded by blurriness, multi-step sampling is preferable to single-step sampling as it progressively recovers finer details. Conversely, in the second example with severe noise, a single sampling step is sufficient to achieve satisfactory results, whereas additional steps may amplify the noise and introduce unwanted artifacts. \textit{(Zoom-in for best view)}}
    \label{fig:multi-step-supp} \vspace{-55mm}
\end{figure*}
\begin{figure*}[t]
    \centering
    \includegraphics[width=\linewidth]{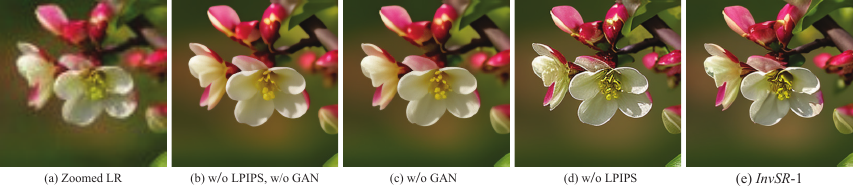}
    \vspace{-6mm}
    \caption{Visual comparisons of the proposed method with various loss configurations. (a) Zoomed LR image, (b) Baseline1 with $\lambda_l=0$ and $\lambda_g=0$, (c) Baseline2 with $\lambda_l=2.0$ and $\lambda_g=0$, (d) Baseline3 with $\lambda_l=0$ and $\lambda_g=0.1$, (e) recommended settings of $\lambda_l=2.0$ and $\lambda_g=0.1$. \textit{(Zoom-in for best view)}}
    \label{fig:loss-ablation}
\end{figure*}
\begin{figure*}
    \centering
    \includegraphics[width=\linewidth]{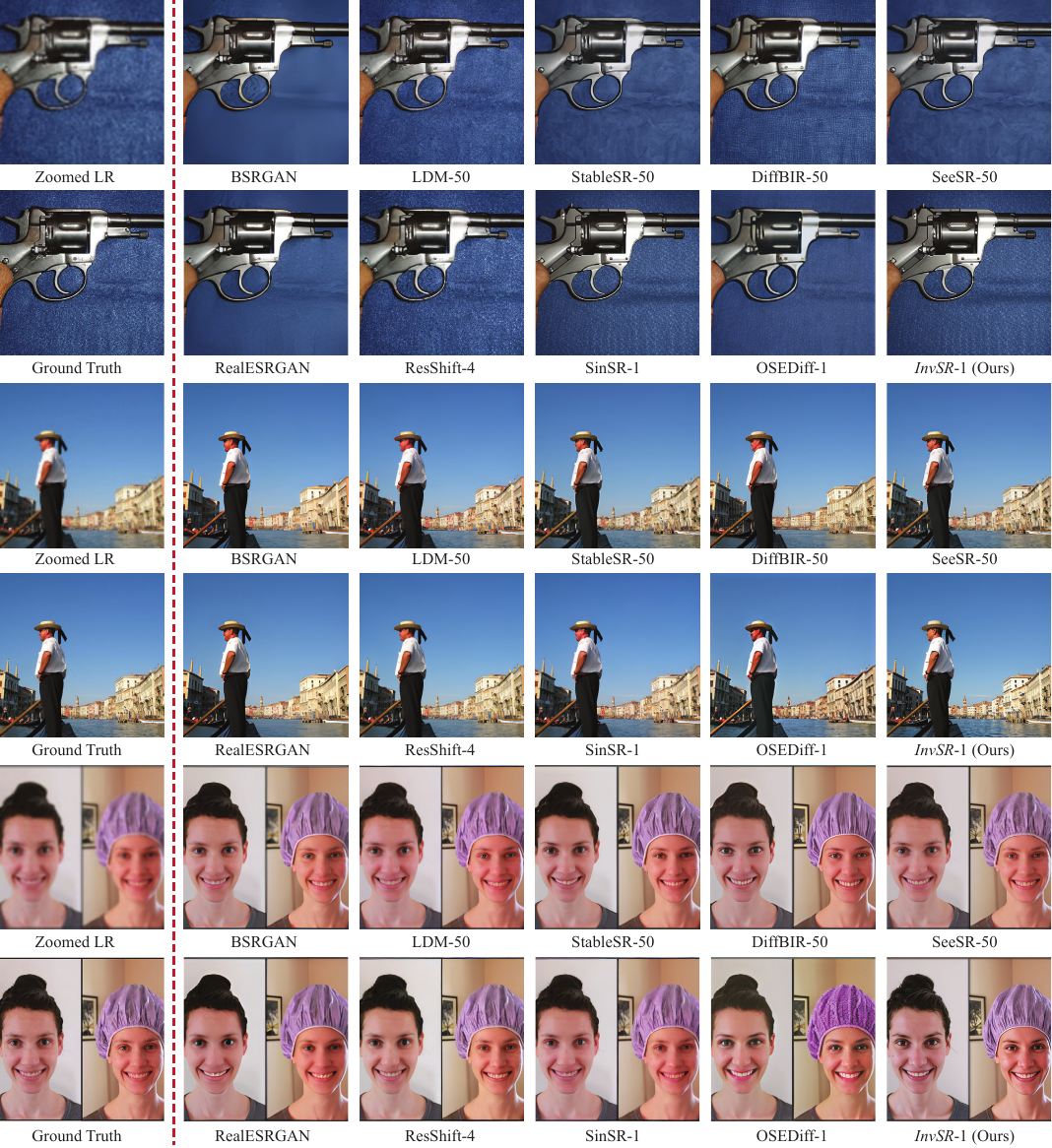}
    \vspace{-6mm}
    \caption{Visual comparisons of various methods on three typical examples from \textit{ImageNet-Test}. For diffusion-based methods, the number of sampling steps is annotated in the format of ``Method name-Steps''.  \textit{(Zoom-in for best view)}}
    \label{fig:imagenet-supp}
\end{figure*}
\begin{figure*}
    \centering
    \includegraphics[width=\linewidth]{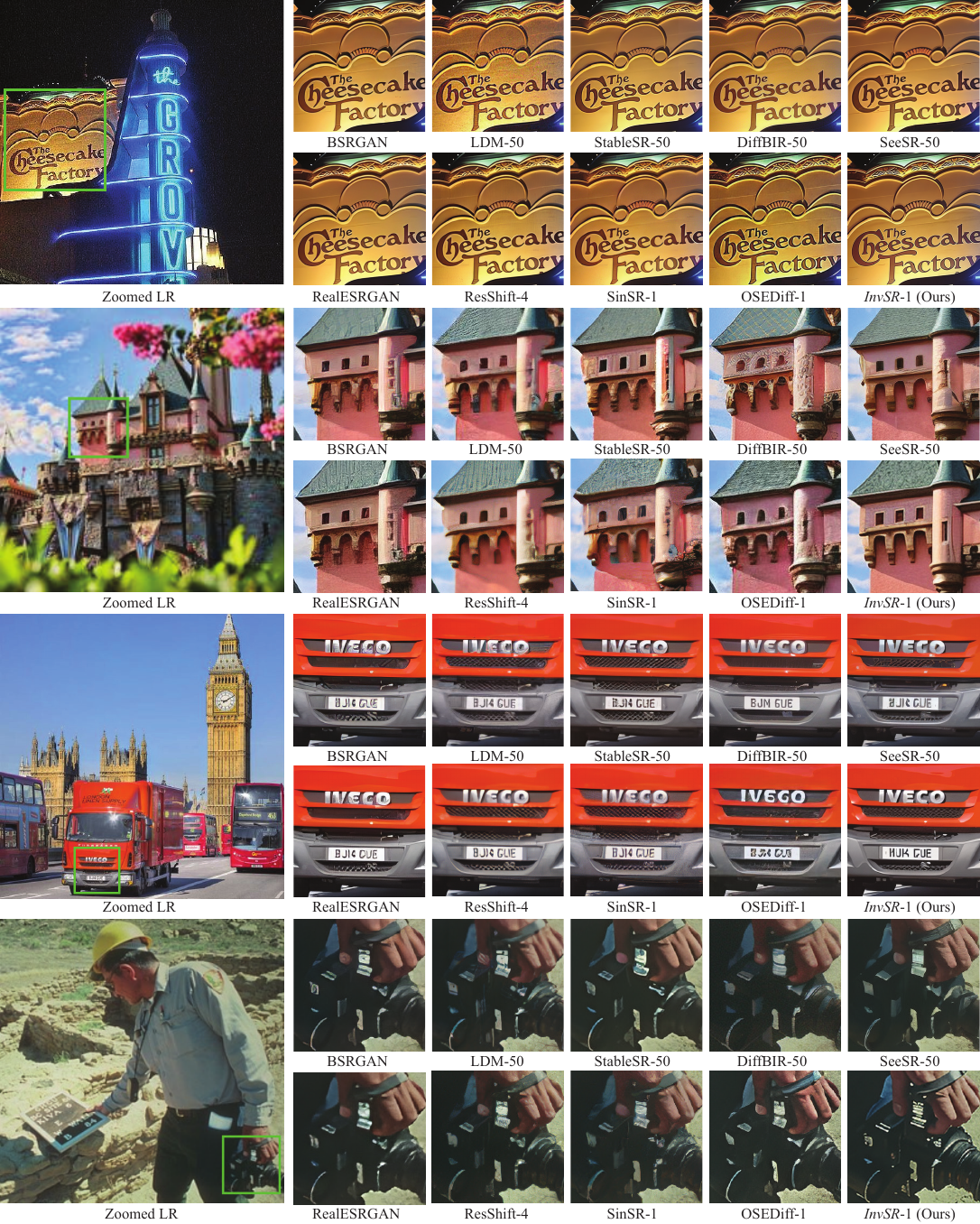}
    \vspace{-6mm}
    \caption{Visual comparisons of various methods on four real-world examples from \textit{RealSet80}. For diffusion-based methods, the number of sampling steps is annotated in the format of ``Method name-Steps''.  \textit{(Zoom-in for best view)}}
    \label{fig:realset80-supp}
\end{figure*}

\end{document}